\pdfoutput=1

\documentclass[11pt]{article}

\usepackage{acl}

\usepackage{times}
\usepackage{latexsym}
\newcommand\scalemath[2]{\scalebox{#1}{\mbox{\ensuremath{\displaystyle #2}}}}

\usepackage[T1]{fontenc}

\usepackage[utf8]{inputenc}

\usepackage{microtype}

\usepackage{helvet}  
\usepackage{courier}  

\usepackage{booktabs}
\usepackage{mathtools}
\usepackage{varwidth}
\usepackage{empheq}
\usepackage{amssymb}
\usepackage{tabularx}
\usepackage{multirow}
\usepackage{epsfig}
\usepackage{amsmath}
\usepackage{cancel}
\usepackage{amsfonts}       
\usepackage{nicefrac}       
\usepackage{bbm}
\usepackage{bm}
\usepackage{subcaption}
\usepackage{defs}
\usepackage{enumitem}

\renewcommand{\vec}[1]{\boldsymbol{#1}}
\renewcommand{\vec}[1]{\boldsymbol{#1}}

\usepackage{siunitx} 
\renewcommand{\vec}[1]{\boldsymbol{#1}}
\usepackage{algorithm}
\usepackage{algpseudocode}

\usepackage{natbib}  
\usepackage{caption} 
\DeclareCaptionStyle{ruled}{labelfont=normalfont,labelsep=colon,strut=off} 
\usepackage{algorithm}

\usepackage{newfloat}
\usepackage{listings}
\floatstyle{ruled}
\newfloat{listing}{tb}{lst}{}
\floatname{listing}{Listing}

\title{Aligned Weight Regularizers for Pruning Pretrained Neural Networks}

\author{James O' Neill \and Sourav Dutta \and Haytham Assem \\
  Huawei Research Center, Dublin, Ireland \\
  \texttt{\{james.o.neil,sourav.dutta2,haytham.assem\}@huawei.com} \\}

\begin{document}
\maketitle

\begin{abstract}
While various avenues of research have been explored for iterative pruning, little is known what effect pruning has on zero-shot test performance and its potential implications on the choice of pruning criteria. This pruning setup is particularly important for cross-lingual models that implicitly learn alignment between language representations during pretraining, which if distorted via pruning, not only leads to poorer performance on language data used for retraining but also on zero-shot languages that are evaluated. 
%
In this work, we show that there is a clear performance discrepancy in magnitude-based pruning when comparing standard supervised learning to the zero-shot setting. From this finding, we propose two weight regularizers that aim to maximize the alignment between units of pruned and unpruned networks to mitigate alignment distortion in pruned cross-lingual models and perform well for both non zero-shot and zero-shot settings. 
We provide experimental results on cross-lingual tasks for the zero-shot setting using XLM-RoBERTa$_{\mathrm{Base}}$, where we also find that pruning has varying degrees of representational degradation depending on the language corresponding to the zero-shot test set. This is also the first study that focuses on cross-lingual language model compression.   
\end{abstract}

\section{Introduction}
Deep neural networks (DNNs) have grown increasingly large in the recent years. This has led to models requiring more storage requirements,  more resources for training and inference (e.g., GPUs and TPUs), longer compute times and larger carbon footprints. This is largely due to the rise of masked self-supervised learning (SSL) which trains DNNs (e.g., Transformers in NLP) on a large collection of samples that do not have task labels but instead use a subset of the inputs as labels. Given the aforementioned challenges, it has become more difficult for machine learning practitioners to use these SSL pretrained models for fine-tuning on downstream tasks. While training tricks such as effective batch sizes, gradient accumulation and dynamic learning rate schedules~\cite{howard-ruder-2018-universal} have improved the efficiency of fine-tuning DNNs under resource constraints, it can still come at a cost, e.g. gradient accumulation leads to less updates. 

Pruning~\cite{lecun1990optimal,reed1993pruning} is a type of model compression method~\cite{bucilua2006model} that aims to address these shortcomings by zeroing out a subset of weights in the DNN, while maintaining performance close to the original model. Retraining is often carried out directly after each pruning step to recover from pruning induced performance drops. This process is referred to as \emph{iterative pruning}. 
Although, iterative pruning has been extensively studied in the SSL setting~\cite{hassibi1993second,han2015compressing,ding2018auto} and the transfer learning setting~\cite{molchanov2016pruning,gordon2020compressing,sanh2020movement}, little is known about pruning DNNs in the zero-shot setting\footnote{Here, zero or one-shot is the conventional usage of the meaning (i.e., number of samples per class), not one-shot pruning~\shortcite{lee2018snip} which is the number of pruning steps used during retraining.} where a model is required to make predictions on a set of samples from classes that are unobserved during training. One salient example is pretrained cross-lingual language models (XLMs)~\cite{conneau2019cross,conneau2020unsupervised} whereby the model is trained with a masked/translation language model (MLM/TLM) objective to predict tokens for a large set of different languages whereby the objective forces the XLM model to learn similar representations for different languages. After cross-lingual pretraining, the model is further fine-tuned to a downstream task in one language (e.g., English) and then evaluated on different languages in the zero-shot setting (e.g., Spanish, French, Chinese, etc.). In this context, applying current pruning methods can damage the XLM cross-lingual alignment that has been learned during pretraining. Ideally, we would aim to prune XLMs in such a way that avoids this alignment distortion which effects the zero-shot performance of pruned XLMs. Additionally, overfitting to the language used for fine-tuning becomes more of an issue due to the progressive reduction in parameters throughout iterative pruning as the remaining weights are relatively large, moving away from an ``aligned'' XLM state. 

This is an important problem to address as the application of large pretrained models in the zero shot-setting for natural language and other modalities (e.g images and audio) is of practical importance e.g., using XLMs in production for multiple languages by only requiring annotations in a single language for fine-tuning, making predictions on unseen classes at test time from pretrained visual representations~\cite{bucher2017generating} using only semantic descriptions (i.e., label similarity to known classes) or zero-shot predictions in pretrained multi-modal models such as CLIP~\cite{radford2021learning}.

Hence, this work addresses the \emph{alignment distortion} pruning problem by introducing {\em AlignReg}, a class of weight regularizers for magnitude-based pruning that force pruned models to have parameters that point in a similar direction or have a similar distribution to the parameters of the original pretrained network. To our knowledge, this is the first study on how iteratively pruned models perform in the zero-shot setting and how the solution differs from solutions found in the non-zero shot setting. We believe our findings have a strong practical implication as well-established pruning criteria may not be suitable given the observed discrepancy between zero-shot performance and the typically reported non-zero shot performance. Moreover, our proposed weight regularizer improves overall pruning generalization in zero-shot cross-lingual transfer. Below, we summarize our \textbf{contributions}. 


\begin{itemize}
\itemsep0em
    \item The first analysis of pruning cross-lingual models, how this effects zero-shot cross-lingual transfer and performance differences to pruning in the SSL setup.
    \item A weight regularizer that mitigates alignment distortion by minimizing the {\em layer-wise Frobenius norm or unit similarity} between the pruned model and unpruned model, avoiding overfitting to single language task fine-tuning. 
    \item A post-analysis of weight distributions after pruning and how they differ across module types in Transformers.  
\end{itemize}

\section{Related Work}

Below we describe regularization-based pruning, other non-magnitude based pruning and how masked language modeling (MLM) implicitly learns to align cross-lingual representations. 

\vspace{.5em}
\textbf{Regularization-based pruning}.
Pruning can be achieved by using a weight regularizer that encourages network sparsity. Three well-established regularizers are $L_0$~\cite{louizos2018learning}, $L_1$ regularization ~\cite{liu2017learning,ye2018rethinking} and the commonly used $L_2$ regularization for weight sparsity ~\cite{han2015learning,han2015compressing}.
~\citeauthor{wang2020neural} have proposed an $L_2$ regularizer that increases in influence throughout retraining and shows the increasing regularization improves pruning performance.
For structured pruning where whole blocks of weights are removed, Group-wise Brain Damage~\cite{lebedev2016fast} and SSL~\cite{wen2016learning} propose to use Group LASSO~\cite{yuan2006model} to learn structured solutions.

\vspace{.5em}
\textbf{Importance-based pruning.}
Magnitude-based pruning (MBP) relies on the assumption that weight or gradient magnitudes have correlation with its importance to the overall output of the network.~\citeauthor{mozer1989skeletonization} instead use a learnable gating mechanism that approximates layer importance, finding that weight magnitudes reflect importance statistics. 
To measure weight importance as the difference in loss between pruned and unpruned network,~\citeauthor{lecun1990optimal} approximate this difference with a Taylor series up to the second order. This involves the product of the gradient and weight magnitude in the 1st term and an approximation of the Hessian and the square of the weight magnitude for the second term. 
However, computing the Hessian and even its approximations~\cite{lecun1990optimal,hassibi1993second,dong2017learning,wang2019eigendamage,singh2020woodfisher} can significantly slow down retraining. 
In our work, we avoid the requirement of computing the Hessian or approximations thereof, as it is not scalable for models such as XLM-R~\cite{conneau2020unsupervised}. 
~\citeauthor{park2019lookahead} have extended MBP to block approximations to avoid pruning lowest weight magnitudes that may be connected to weights in adjacent layers that have high weight magnitude.~\citeauthor{lee2020layer} have provided a method to automatically choose the sparsity of layers by using the rescaled version of weight magnitude to incorporates the model-level distortion incurred by pruning.

\paragraph{Implicit Alignment in Pretrained MLMs}
In context of multi-task learning,~\newcite{chen-etal-2020-recall} minimize the mean squared error between pretrained weights and weights being learned for a set of different source tasks to avoid catastrophic forgetting in the continual learning setting. 
~\newcite{conneau-etal-2020-emerging} have found that multilingual MLM (i.e training with an MLM objective with concatenated text for multiple languages) naturally leads to models with strong cross-lingual transfer capabilities. Additionally, they find that this is also found for monolingual models that do not share vocabulary across monolingual corpora and the only requirement is that weight sharing is used in the top layers of the multi-lingual encoder. In the context of our work, we want to bias our fine-tuned and iteratively pruned model to have similar geometric properties and symmetries to these pretrained MLMs to preserve zero-shot cross-lingual transfer.


\section{Methodology}
In this section, we describe how our proposed {\em AlignReg} weight regularizers can improve pruning performance in both supervised learning and zero-shot pruning settings. We focus on two regularizers, namely, {\em a neuron correlation-based regularizer} ($\mathrm{cosine}$-$\mathrm{MBP}$) and {\em Frobenius layer-norm regularizer} ($\mathrm{frobenius}$-$\mathrm{MBP}$). 

Let $\mathcal{D}: = \{X_i, y_i\}^D_{i=1}$ where each $X_i$ of $D$ training samples consists of a sequence of vectors $X_i := (\vec{x}_1, \ldots, \vec{x}_n)$ and $\vec{x}_i \in \mathbb{R}^{d}$ (e.g., $d=512$). For structured prediction (e.g., NER and POS), $y_i \in \mathbb{R}^{n \times c}$ and for single and pairwise sentence classification, $y_i \in \mathbb{R}^{c}$ where $c$ is the number of classes.
Let $\theta = (\theta_1, \ldots, \theta_L)$ be the parameters of a pretrained network $f$ with $L$ layers, where $\theta_l$ refers to the parameters, including weight matrix $\mat{W}_l$ and bias $b_l$, at layer $l$. Let $f_{\tilde{\theta}}$ be a
network with parameters $\tilde{\mathbf{\theta}}$ consisting of weights $\tilde{\mat{W}}_l \in \mathbb{R}^{N_{l-1}\times N_l}$ and bias $\vec{\tilde{b}}_l \in \mathbb{R}^{N_l}$ where $N_l$ is the number of units in the $l$-th layer. Here, $\tilde{\mathbf{W}}_l := \mat{W}_l \mat{M}_l$ where $\mat{M}$ is the pruned mask. 
%
For MBP~\cite{karnin1990simple} we remove weights of $\mat{W}_l$, $\forall l \in L$ with the smallest absolute weight magnitude until a specified percentage $p$ of weights are removed. Note that this is a layer-wise process and requires the pruned weights to be masked with $\mat{M}_l$ which has 0 entries corresponding to weights to be pruned and 1 entries for unpruned weights $\mat{W}_l$.  Global MBP can also be used whereby the weights $\{\mat{W}_l\}^{L}_{l=1}$ are first vectorized and concatenated prior to choosing $p$ lowest weight magnitudes. Unlike layer-wise MBP, the percentage of weights removed in each layer can vary for global-MBP. Typically, weight regularization is used with MBP to encourage weight sparsity. Thus the objective for iterative pruning can be expressed as,
\begin{equation}
    \mathcal{L}_{\theta} :=\frac{1}{D} \sum^D_{i=1} \ell_{ce}\big(f_{\tilde{\mathbf{\theta}}}(\mat{X}_i), \vec{y}_i\big) + \lambda ||\tilde{\mathbf{\theta}}||_0 
\end{equation}

where $\lambda$ controls the influence of the weight magnitude regularization. We now describe our proposed {\em AlignReg}.


\subsection{\emph{AlignReg} - Pruning-Aware Regularization}
{\em AlignReg} can be used to align weights unit-wise or layer-wise between unpruned and pruned networks. We initially discuss the $\mathrm{cosine}$-$\mathrm{MBP}$ regularizer. 

$\mathrm{cosine}$-$\mathrm{MBP}$ aims to preserve the inherent cross-lingual alignment, during iterative  pruning, by minimizing the angle between parameter vectors of the same unit in the pruned and unpruned network. The intuition is that cross-lingual alignment relies more on parameter vector direction than vector magnitudes. Moreover, as the network is being pruned, the weights will consequently change weight magnitude to account for the information loss. 
To apply \emph{AlignReg} to linear layers within Transformers, we compute the pairwise cosine similarity between pairs of pruned weights $\mathbf{\tilde{W}}_l \subset \tilde{f}$ and unpruned weights $\mat{W} \subset f$ for all $l$-th layers. For $\mat{W}_l \in \mathbb{R}^{N_{l-1}\times N_l}$ of the $l$-th layer, the average weight correlation is
\vspace{-.75em}
\begin{equation}\label{eq:cos_1}
\begin{gathered}
\rho(\tilde{\mathbf{W}}_l,\mat{W}_l) =  
\frac{1}{N_l} \sum_{i=1}^{N_l}\frac{|\mat{W}^{\top}_{li} \tilde{\mathbf{W}}_{li}|}{||\mat{W}_{li}||_2||\tilde{\mathbf{W}}_{li}||_2}
\end{gathered}
\end{equation}

where $\mat{W}_{li}$ is $i$-th column of the matrix corresponding to the $i$-th unit of the $l$-th layer. Intuitively, $\rho(\mat{W}_l, \mathbf{\tilde{W}}_l)$ is the average cosine similarity between weight vectors of the same unit at the $l$-th layer of the pruned and unpruned network.
Adding {\em AlignReg} to the objective results in Equation \eqref{eq:total_loss}, 
\vspace{-.75em}
\begin{equation}\label{eq:total_loss}
    \mathcal{L}_{\theta} := \ell_{ce}\big(f_{\tilde{\theta}}(\mat{X}), \vec{y}\big) - \frac{\lambda}{L} \sum_{l}^{L}\rho\big(\tilde{\mathbf{W}}_l, \mat{W}_l\big)
\end{equation}

where $\lambda \in [0, \infty)$ controls the importance of AlignReg relative to the main cross-entropy loss $\ell_{ce}(\cdot, \cdot)$.
The gradient of the loss w.r.t to $\theta$ is then expressed as equation \eqref{eq:grad}, 
\begin{equation}\label{eq:grad}
\scalemath{0.875}{
\nabla_{\theta} \mathcal{L}_{\theta} := \nabla_{\tilde{\theta}} \ell_{ce}(f_{\tilde{\theta}}(\mat{X}), \vec{y}) - \frac{\lambda}{L} \sum_{l}^{L}\frac{\partial \rho\big(\mathbf{\tilde{W}}_l, \mat{W}_l\big)}{\partial \tilde{\mathbf{W}}_l}    
}
\end{equation}


where $\frac{\partial \rho(\mathbf{\tilde{W}}_l, \mat{W}_l)}{\partial \tilde{\mathbf{W}}_l}$ is a function of the `2-norm of the matrices in $\mat{W}_l$. For the element $\mat{W}_{l,(i,j)}$ of $i$-th row and $j$-th column in $\mat{W}_l$, we have
\vspace{-.5em}
\begin{equation}\label{eq:grad_2}
\begin{gathered}
\scalemath{0.85}{
\frac{\partial \rho(\mathbf{\tilde{W}}_l, \mat{W}_l)}{\partial \mathbf{\tilde{W}}_{l,(i,j)}} = \frac{1}{N_l-1}
\sum_{j=1}^{N_l}\Big(\sign(\mat{W}^{\top}_{l,(,j)} \mathbf{\tilde{W}}_{l,(,j))}} \\
\scalemath{0.85}{
\Big[\frac{\mathbf{\tilde{W}}_{l,(i,j)}}{||\mat{W}_{l,(,j)}||_2 ||\mathbf{\tilde{W}}_{l,(,j)}||_2} \scalebox{1.75}[1.0]{-}
\frac{\mat{W}_{l,(i,j)} \mat{W}^{\top}_{l,(,j)} \mathbf{\tilde{W}}_{l,(,j)}}{||\mat{W}_{l,(,j)}||^3_2 ||\mathbf{\tilde{W}}_{l,(,j)}||_2}
\Big]\Big)
}
\end{gathered}
\end{equation}

where $\mat{W}_{l,(,j)}$ and $\tilde{\mathbf{W}}_{l,(,j)}$ are $j$-th column in $\mat{W}_l$ and $\tilde{\mathbf{W}}_l$, respectively.
Thus, this regularization favors solutions with high cosine similarity between units of pruned and unpruned networks.
%
We also consider a layer-wise $\rho(\mat{W}, \tilde{\mathbf{W}})$ that relaxes the unit-level alignment to whole layers. This is partially motivated due to the fact neural networks can exhibit similar output activation behavior even when neuron weights have been permuted within the layer~\cite{brea2019weight}. To perform this we simply apply Equation \eqref{eq:cos_1} with vectorized weights $\rho(\mathrm{vec}(\tilde{\mathbf{W}}_l),\mathrm{vec}(\mat{W}_l))$ and the subsequent partial derivatives in Equations \eqref{eq:grad} and \eqref{eq:grad_2} are applied for updating $\tilde{\mathbf{W}}_l$. In our experiments we did not see a significant difference using vectorized weights and thus use unit-wise cosine similarity.

Algorithm \ref{alg:alignreg} shows how AlignReg is applied for a single mini-batch update during an iterative pruning epoch. 

\begin{algorithm}[t]\caption{AlignReg Pruning}\label{alg:alignreg}
\begin{algorithmic}[1]
\State \textbf{Input:} Weight tensors $\mat{W}_1,\dots,\mat{W}_L$ of a fine-tuned network, $p$ percentage of weights to remove per layer
\State \textbf{Output:} Pruned weight tensors $\tilde{\mathbf{W}}_1,\dots \tilde{\mathbf{W}}_L$
\For{$l=1,\dots,L$}
\State Compute $\rho(\tilde{\mathbf{W}}_l, \mathbf{W}_l)$ with \textbf{Eq.2}.
\State Set $\tilde{\mathbf{W}}_{s_i}$ as $s_l$-th smallest element of $\tilde{\mathbf{W}}$
\State Set $\mathbf{M}_l\leftarrow\mathbbm{1}\{\mathbf{W}_l-\tilde w_{s_l}\ge 0\}$
\State Set $\tilde{\mathbf{W}}_l \leftarrow \mathbf{M}_l\odot \mathbf{W}_l$
\EndFor
\State Compute $\mathcal{L}_{\theta}$ according to \textbf{Eq.3}

\end{algorithmic}
\end{algorithm}

\paragraph{Relaxing Unit-Wise AlignReg To A Layer-Wise Frobenius Distortion Formulation}
Thus far we have described the application of cosine similarity as a measure of similarity between unpruned and pruned weights of the same units. However, this may be a strict constraint, particularly at high compression rates where the remaining weights for a unit are forced to have higher norms to allow zeroed weights. Hence, an alternative measure is the layer-wise Frobenius norm (Frobenius-MBP) regularizer based on the difference between weights $||\mat{W} - \tilde{\mathbf{W}}||_F$. 
MBP itself can be viewed in terms of minimizing the Frobenius distortion~\cite{han2015compressing,dong2017learning} as $\min_{\mat{M}:||\mat{M}||_0=p} ||\mat{W} - \mat{M} \odot \mat{W} ||_F$ where $\odot$ is the Hadamard product, $||\cdot||_0$ denotes the entrywise 0-norm, and $p$ is a constraint of the number of weights to remove as a percentage of the total number of weights for that layer. 
In the zero-shot setting, we need to account for out-of-distribution Frobenius distortions, such as \emph{alignment distortion} in XLM due to pruning and overfitting to a single language. Taking the view of Frobenius distortion minimization when using our weight regularizer, we reformulate it to include Frobenius-MBP as, 
\begin{equation}
\scalemath{0.875}{
\min_{\mat{M}:||\mat{M}||_0=p} \Big[||\mat{W} \text{-} \mat{M} \odot \mat{W} ||^2_F + \lambda||\mat{W}^{T} \text{-} \mat{M} \odot \mat{W} ||^2_F \Big] 
}
\end{equation}

where $\mat{W}^{T}$ are the weights from the pretrained model prior to fine-tuning that is cross-lingually aligned from the masked language modeling (MLM) pretraining objective. In our experiments, $\lambda = 5 \times 10^{-4}$. 
\paragraph{\emph{frobenius}-MBP Implicitly Aligns Eigenvectors} To explicitly show that the Frobenius distortion minimization aligns fine-pruned and pretrained parameter vectors we expect their eigenvectors to also be close. We can use the Eckart-Young-Mirsky Theorem~\citep{GOLUB1987317} to express Frobenius distortion minimization as~\autoref{eq:eigen},
\begin{equation}\label{eq:eigen}
||\mat{W}^T - \mat{M} \odot \mat{W}||^2_F = ||\vec{\Sigma} - \mat{U}^{\top}\mat{M}\odot\mat{W}\mat{V}||^2_F    
\end{equation}
 where the unitary invariance under the 2-norm that $\mat{U}$,$\mat{V}$ vanishes and singular value matrix is left to approximate $\mat{W}^T$, hence the inclusion of $\vec{\Sigma}$. We express $\mat{X}=\mat{U}_k\Sigma^{12}_k$, $\mat{Y}=\Sigma^{12}_k\mat{V}^{\top}_k$ and $\mat{X}\mat{Y} = \mat{A}_k$. Hence, we can further describe the minimization as $||\Sigma - \mat{U}^{\top}\mat{W}^T_k\mat{V}||^2_F$ and since $\mat{X}$, $\mat{Y}$ are unitary, $||\Sigma - \Sigma_k||^2_F$.

\begin{figure*}[ht]
    \centering
    \includegraphics[width=1.\linewidth]{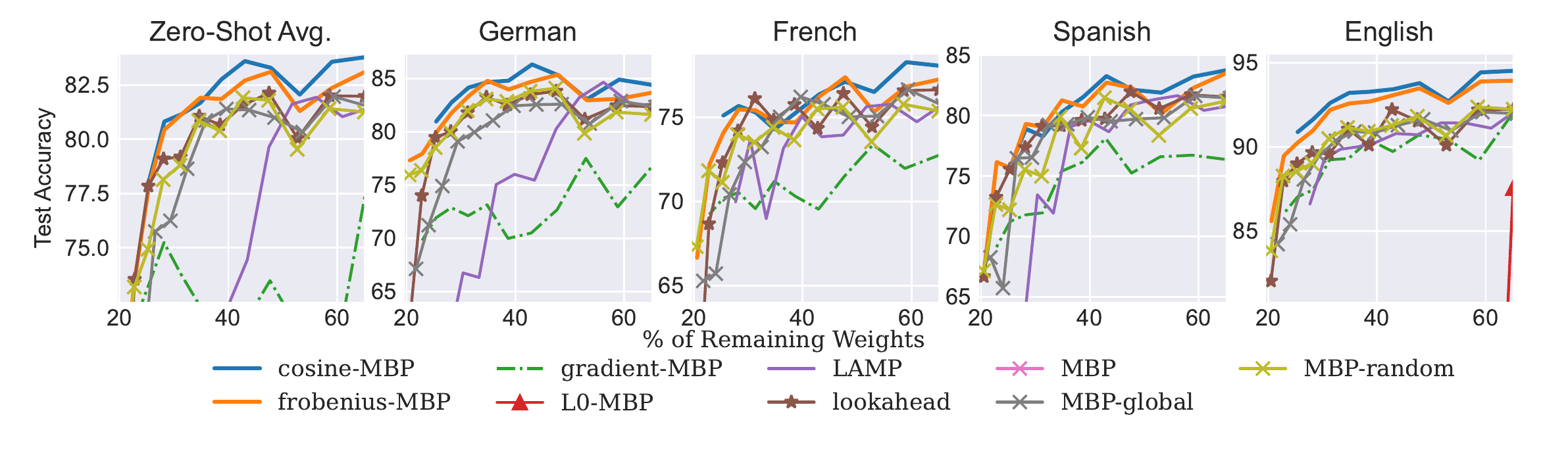}
    \caption{\textbf{English and Zero-Shot Test Accuracy on News Classification. }}
    \label{fig:nc}
    \vspace{-1em}
\end{figure*}

\subsection{Connections to Knowledge Distillation}
Knowledge distillation (KD) works by using outputs of the last layer~\cite{hinton2015distilling} or intermediate layers~\cite{romero2014fitnets} as additional soft targets. {\em AlignReg} regularizers instead operate directly on minimizing a divergence or distance between weight tensors as opposed to their corresponding output activations. Hence, {\em AlignReg} does not necessarily need training data as it operates directly on aligning weight tensors. Since the networks that are used for alignment are architecturally identical, we can show that maximizing weight similarity is equivalent to minimizing distance between their corresponding output activations~\cite{romero2014fitnets} when the norm of input $Z$ is smaller than the output range of $\sigma$. For our experiments, we use XLM-RoBERTa$_{\mathrm{Base}}$ which contain Gaussian Linear Error Unit (GeLU) activation functions, which can be formulated as $\sigma(\mat{Z}_{li}) := \mat{Z}_{li}/2 (1.0 + \mathrm{erf}(\mat{Z}_{li}/ \sqrt{2.0}))$ where $\mathrm{erf}$ is an error function, $\sigma(\cdot)$ is a monotonic activation function and $\vec{Z}_{li}$ is the input vector. The GELU activation has the properties that for $\mat{Z}_{li} > 0$ it is equivalent to the ReLU activation and $\mat{Z}_{li} \leq 0$ it tends to -1. For $\mat{Z}_{li} > 0$, $||\mat{Z}_{li}||_2 \leq 1$ and a monotonic piecewise linear function $\sigma(\cdot)$, the inequality in~\autoref{eq:inequality} holds.

\begin{equation}\label{eq:inequality}
\begin{aligned}
& || \mat{W}_{li} \text{-} \mat{M}_{li} \odot \mat{W}_{li}||_F \leq \\ 
& || \sigma(\mat{Z}_{l}\mat{W}_{li}) - \sigma (\mat{Z}_{li} \mat{M}_{li} \odot\mat{W}_{li})||_F  
\end{aligned}
\end{equation}

Layer normalization leads to features having zero mean and unit variance and hence  $||\mat{Z}_{li}||_2 \leq 1$.
Hence, minimizing the Frobenius distortion of pruned and unpruned weights is equivalent to minimizing the mean squared error (MSE) between output activations, as is the knowledge distillation method used for FitNets~\cite{romero2014fitnets}. In contrast, KD using FitNets encourages the student network to have activation outputs that are the same as the teacher with permutation invariance on the units incoming weights, not restricting the weights to be similar. Unlike KD, this minimization can be performance without any data. 

\section{Experimental Setup}
\paragraph{Datasets.}
We perform experiments on multilingual tasks from the XGLUE benchmark~\cite{liang2020xglue} with pretrained XLM-R$_{\mathrm{Base}}$. This covers pairwise classification (QAM, QADSM, WPR, XNLI), sentence classification (NC) and structured prediction (NER and POS) tasks.  
\paragraph{Iterative Pruning Details.} Texts are tokenized using the SentencePiece BPE tokenizer~\cite{sennrich2015neural} with a vocabulary of 250K tokens. For structured prediction tasks (POS and NER), a single layer feed-forward (SLFF) token-level classifier is used on top of XLM-R$_{\mathrm{Base}}$ and for sentence-level task a SLFF sentence-level classifier is used. The batch size is 32, the learning rate is $5\cdot10^{-6}$ and the maximum sequence length is set to 256 for all tasks, except for POS in which we use a learning rate of $2\cdot10^{-5}$ with the $\mathtt{adam}$ optimizer~\cite{kingma2014adam} with weight decay (AdamW) and a max sequence length of 128. We carry out a pruning step after each 15 training epochs, uniformly pruning 10\% of the parameters at each pruning step. We omit the pruning of embedding layers, layer normalization parameters and the classification layer as they account for a relatively small number of the total parameter count ($<1$\%) and play an important role in XLM generalization. 
Although prior work has suggested non-uniform pruning schedules (e.g., cubic schedule~\cite{zhu2017prune}), we did not see major differences to uniform pruning in preliminary experiments. Each task is trained with English data only and evaluated on all available languages for that task. Hence, we expect the percentage of achievable compression to be lower as performance in the zero-shot cross-lingual setting to be more difficult than the monolingual setting (e.g., GLUE tasks).

\begin{figure*}[ht]
    \centering
    \includegraphics[width=1.0\linewidth]{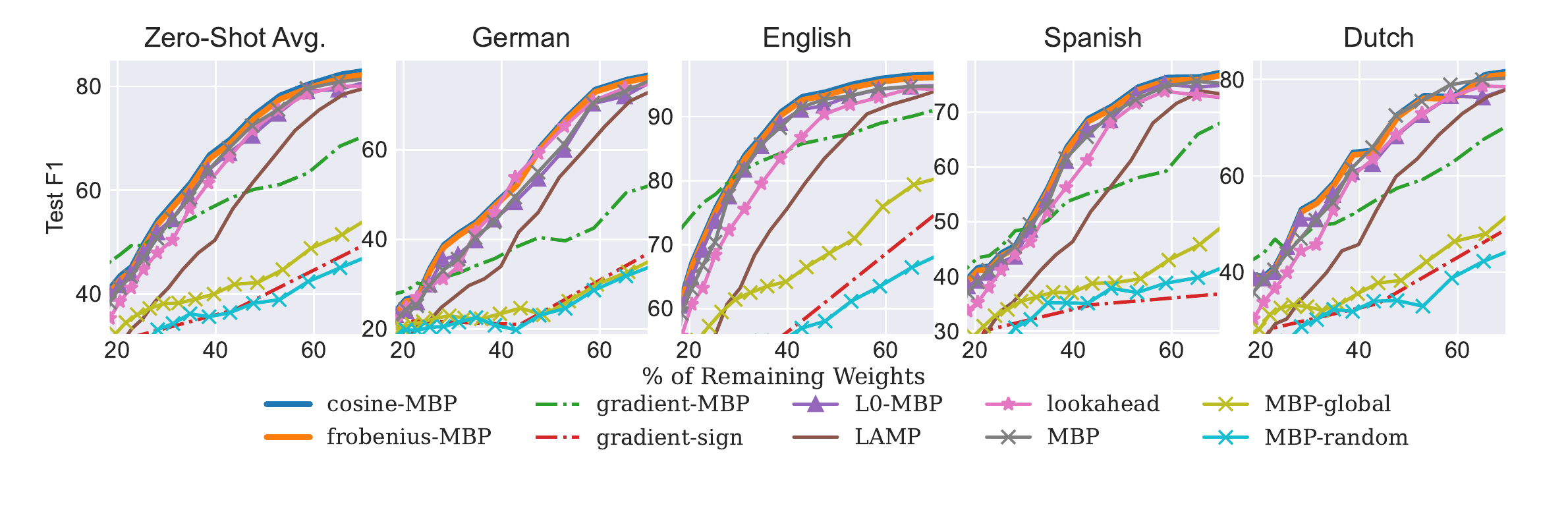}
    \vspace*{-10mm}
    \caption{\textbf{Zero-Shot Test F1 on Named Entity Recognition. }}
    \label{fig:ner}
\end{figure*}
\begin{figure}[ht]
\vspace{-1em}
    \centering
    \includegraphics[width=1.\linewidth]{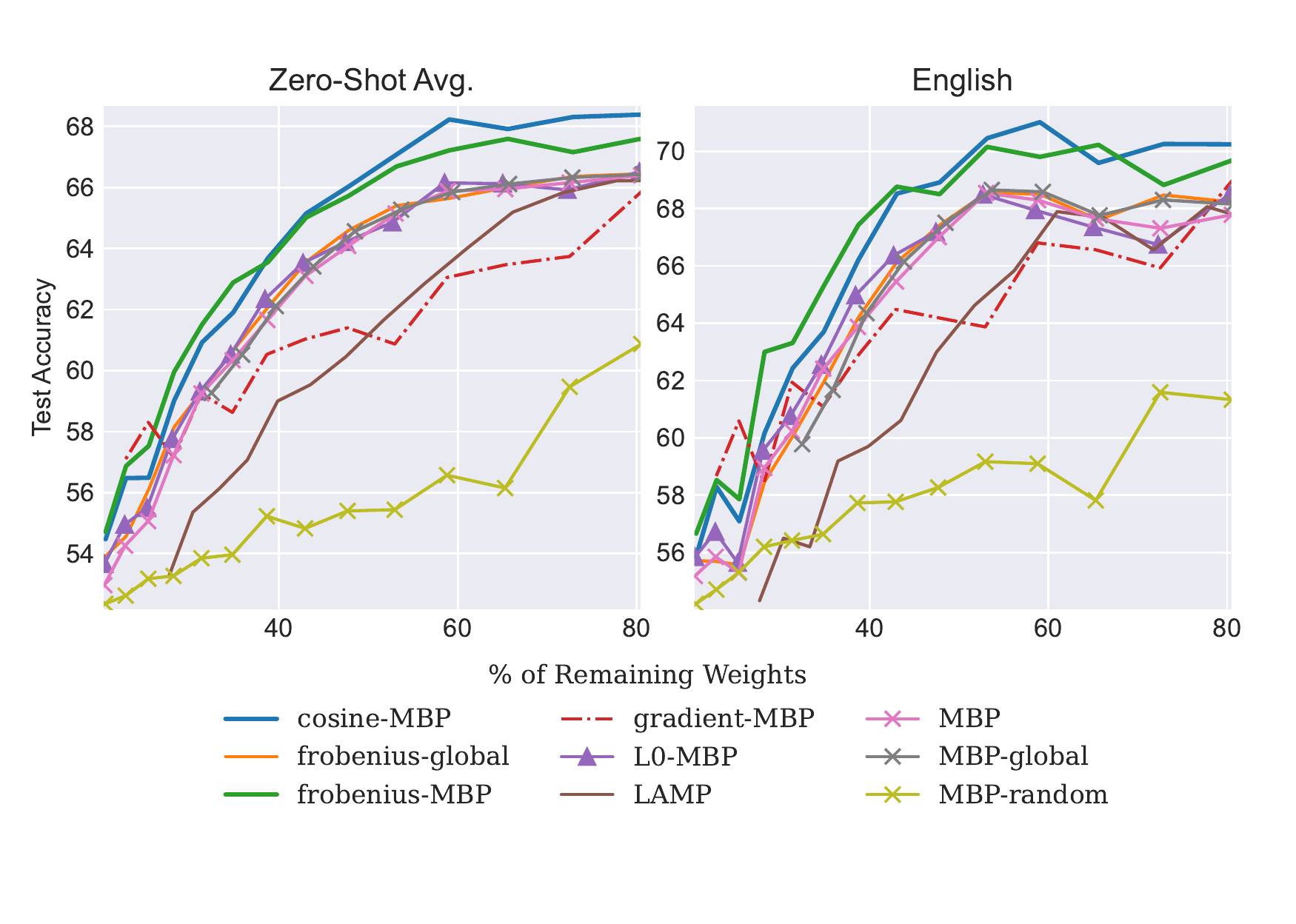}
    \vspace*{-10mm}
    \caption{\textbf{Question Answer Matching Test Accuracy.}}
    \label{fig:en_vs_avg_qam}
\end{figure}

\paragraph{Pruning Baselines.}
Below lists our pruning baselines. 
\textbf{Random Pruning}~\shortcite{stepniewski1997pruning} - weights are pruned uniformly at random across all layers to a chosen fraction.\textbf{Layer-wise Magnitude Pruning} (MBP)~\cite{janowsky1989pruning,mozer1989skeletonization} - for each layer, weights with the lowest absolute value (LAV) are pruned.
\textbf{Layer-wise Gradient Magnitude Pruning}~\cite{sun2017meprop} - for each layer, prunes the weights with LAV of the accumulated gradients evaluated on a batch of inputs.
\textbf{Global Magnitude Pruning} (Global-MBP)~\cite{karnin1990simple} - prunes weights with LAV anywhere in the DNN.
$L_0$ \textbf{norm MBP}~\cite{louizos2018learning} - uses non-negative stochastic gates that choose which weights are set to zero as a smooth approximation to the non-differentiable $L_0$-norm. 
\textbf{Lookahead pruning (LAP)}~\cite{park2019lookahead} - prunes paths that have smallest weight magnitude across blocks of layers, unlike MBP which treats layers independently. 
\textbf{Layer-Adaptive Magnitude Pruning (LAMP)}~\cite{lee2020layer} adaptively sets the pruning ratio of each layer. 
\begin{figure*}
    \centering
    \includegraphics[width=1.\linewidth]{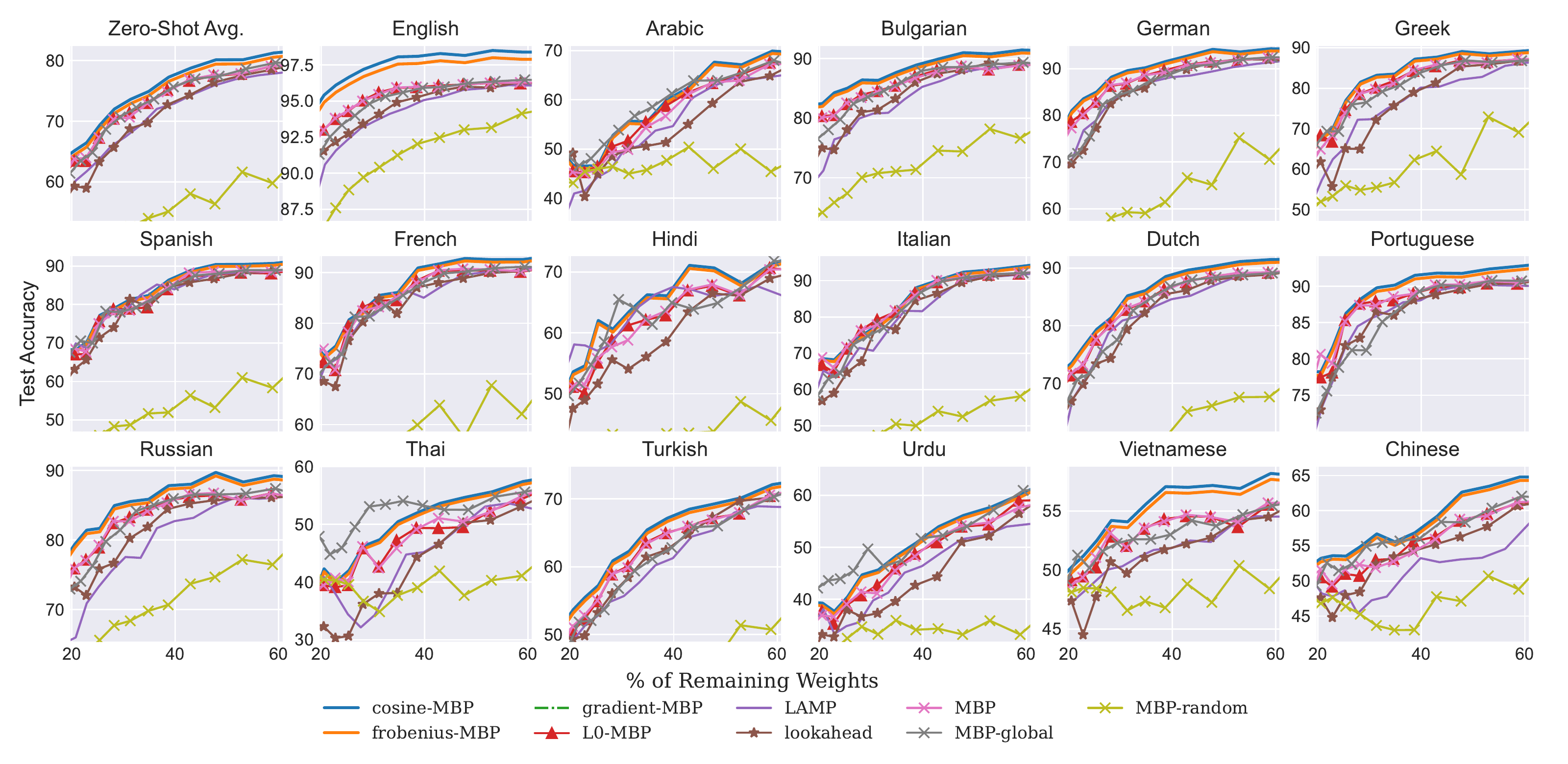}
    \vspace*{-7mm}
    \caption{\textbf{Part of Speech Tagging Test Accuracy.}}
    \label{fig:en_vs_avg_pos}
    \vspace*{-5mm}
\end{figure*}

\section{Empirical Results}
We now discuss results on the XGLUE tasks.
\paragraph{News Classification (NC)}
Figure~\ref{fig:nc} shows the results on news classification where a category for news article is predicted and evaluated in 5 languages and trained and iteratively pruned on English text. Firstly, we observe the trend in iterative pruning performance degradation is somewhat volatile. From preliminary experiments we found news classification to require only 3 epochs to converge for standard fine-tuning on XLM-RoBERTa$_{\mathrm{Base}}$. We find that this task is relatively ``similar'' to the pretraining task and therefore able to easier recover from pruning steps. Overall, both Cosine-MBP and Frobenius-MBP consistently lead to the best zero-shot test performance across both pruning steps and languages. 

\begin{figure}[ht]
    \centering
    \includegraphics[scale=0.5]{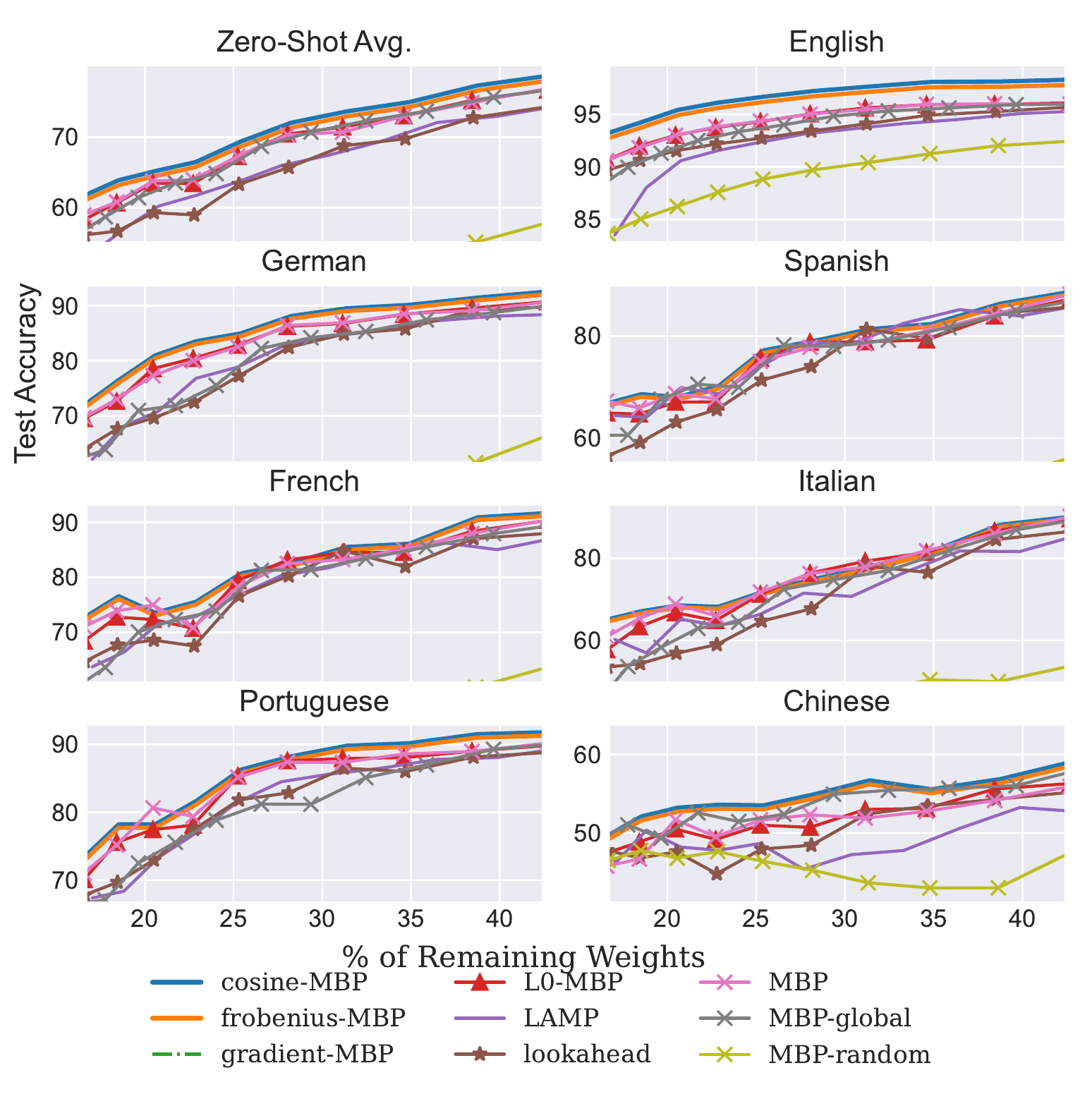}
    \caption{\textbf{Web-Page Ranking Test Matthew's Correlation Coefficient. }}
    \label{fig:wpr_results}
    \vspace{-5mm}
\end{figure}

\paragraph{Question Answer Matching (QAM)}

Figure~\ref{fig:en_vs_avg_qam} shows the test accuracy on English and the zero-shot test accuracy on French and German for Question-Answer Matching (QAM). This involves predicting whether a question is answered correctly or not given a question-answer pair. We find that Frobenius-MBP and Cosine-MBP maintain higher accuracy across multiple pruning steps, outperforming baselines. More generally, we see there is close to 2\% drop in average test accuracy drop in French and German when compared to testing on samples from the same language used in training.  

\begin{figure*}[ht]
  \centering
  \includegraphics[width=1.\linewidth]{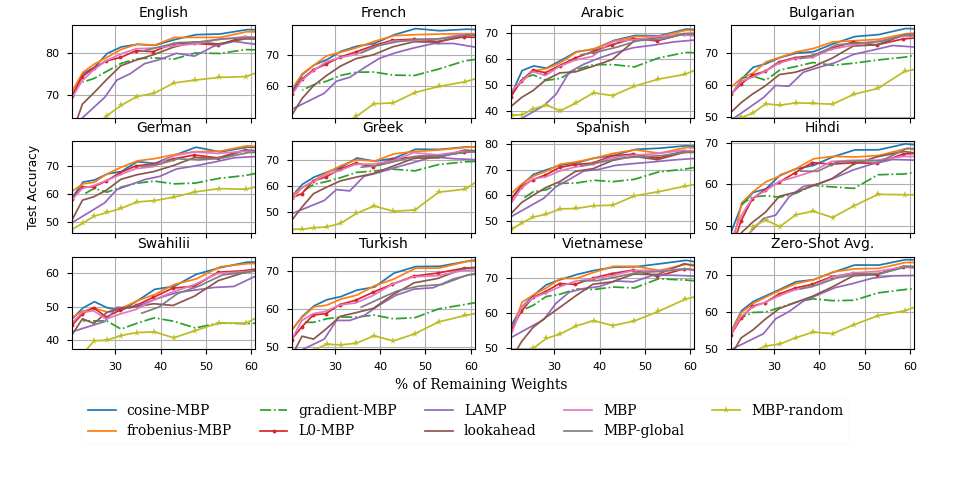}
 \vspace{-13mm}
\caption{\textbf{Zero-Shot XNLI Results Per Language After Iteratively Fine-Pruning XLM-RoBERTa$_\mathrm{Base}$}}
\label{fig:xnli_perf}
\end{figure*}

\paragraph{Named Entity Recognition (NER)}
The Named Entity Recognition (NER) cross-lingual dataset is made up of CoNLL-2002 NER and
CoNLL-2003 NER~\cite{sang2003introduction}, covering  English, Dutch, German and Spanish with 4 named entities. From Figure~\ref{fig:ner} we find that cross-lingual transfer of pruned models is most difficult in German and Dutch, which both come from the same language family, sharing commonalities such as word order and having similar vocabularies. The primary reason for the difficulty in maintaining performance in high compression rates for this NER dataset is that there is only 15k training samples, being significantly lower than the remaining XGLUE tasks (the majority contains 100k training samples). Thus, not only is there less training data to recover directly after each pruning step, but the pruning step interval itself is shorter. In contrast, English test performance is close to the original performance up until 25\% of remaining weights, unlike the remaining languages. We find that gradient-MBP eventually overtakes MBP approaches past 20\% remaining weights.  However accuracy has reduced too much at this compression level. We find that Cosine-MBP and Frobenius-MBP weight regularizers achieve the best performing pruned model performance above 20\% remaining weights, with Lookahead pruning and $L_0$ regularized MBP being competitive in zero-shot performance. 


\paragraph{Part of Speech Tagging (POS)}

The Part of Speech (PoS) tagging dataset consists of a subset of the Universal Dependencies treebank~\cite{nivre2020universal} and covers 18 languages. In Figure ~\ref{fig:en_vs_avg_pos}, we see both Cosine-MBP and Frobenius-MBP tend to outperform baselines, although $L_0$-based pruning~\cite{louizos2018learning} has similar performance to Cosine-MBP for zero-shot accuracy. There is also a clear discrepancy between SSL accuracy (English) versus zero-shot accuracy (Average), the latter following closer to linear decay after 40-50\% of weights remaining. Generally, both Cosine-MBP and Frobenius-MBP outperform baselines with the exception of Thai and Urdu at higher compression rates ($<40$\%), both being some of the most under-resourced languages of all 18 languages.

\paragraph{Web Page Ranking} aims to predict whether a web page is relevant (1-5 ratings, ``bad'' to ``perfect'') to an input query and it is evaluated for 7 languages using the Normalized Discounted Cumulative Gain (nDCG). From~\autoref{fig:wpr_results}, we see that between the 15\% - 45\% region the average zero-shot performance degrades faster than the English language used for training. In contrast, semantically and syntactically different languages from English, such as Chinese, already suffer from loss of alignment due to pruning as the performance gap between proposed methods (and baselines) and random pruning is shortened. 

\paragraph{Cross-Lingual Natural Language Inference (XNLI)}
Figure ~\ref{fig:xnli_perf} shows the zero-shot cross-lingual transfer for various unstructured pruning methods. We find that both the accuracy on the English test (i.e SSL generalization) and the average zero-shot test accuracy are consistently improved using Cosine-MBP and Frobenius-MBP, outperforming $L_0$ pruning, Lookahead pruning and LAMP. We find that morphologically rich languages such as Arabic, Swahili and Turkish degrade in performance linearly once performance begins to drop after 60\% of the remaining weights are pruned. This trend is roughly followed for all MBP-based pruning methods. Additionally, test accuracy on English can be maintained within 10\% accuracy drop of the original test accuracy up to 20\% of remaining weights for MBP, while Swahili can only be within a 10\% accuracy drop up to 55\% of the remaining weights. Hence, iterative pruning in the zero-shot setting leads to faster performance degradation for languages that are typologically or etymologically further from the language used for fine-tuning. 

When comparing, English and the average zero-shot test accuracy we see that the slope is steeper after the inflection point\footnote{The point which the performance slope significantly steepens and drops are relatively large to previous pruning steps.} for all pruning methods, not to mention the greater than 10\% accuracy drop across pruning steps. 


\paragraph{XGLUE Average Result}
Finally, in Table \ref{tab:ft_all} we show the overall and average task \textit{understanding} scores on the XGLUE benchmark for our proposed {\em AlignReg} weight regularizer and the pruning baselines. We find that the use of {\em AlignReg} Cosine-MBP and Frobenius-MBP better preserves cross-lingual alignment during model pruning, thereby outperform other MBP baselines, including LAMP and Lookahead pruning, based on improved zero-shot cross-lingual performance.

\begin{table*}[t]
\begin{center}
    {\small
    \resizebox{.995\linewidth}{!}{
    \begin{tabular}[b]{l|cccccccc|l}
    \toprule[1.05pt]
    \textbf{Prune Method} & \textbf{XNLI} & \textbf{NC} & \textbf{NER} & \textbf{PAWSX} & \textbf{POS} & \textbf{QAM} & \textbf{QADSM} & \textbf{WPR} &  \textbf{Avg.} \\ 
    \midrule

    No Pruning & 73.48 & 80.10 & 82.60 & 89.24 & 80.34 & 68.56 & 68.06 & 73.32 & 76.96 \\
    \midrule
    \midrule
    Random & 51.22 & 70.19 & 38.19 & 57.37 & 52.57 & 53.85 & 52.34 & 70.69 & 55.80  \\
    Global-Random & 50.97 & 69.88 & 38.30 & 56.74 & 53.02 & 54.02 & 53.49 & 69.11 & 55.69  \\
    $L_0$-MBP & 64.75 & 78.98 & 56.22 & 72.09 & 71.38 & 59.31 & 53.35 & 71.70 & 65.97 \\
    $L_2$-MBP & 64.30 & 78.79 & 54.43 & 77.99 & 70.68 & 59.24 & 60.33 & 71.52 & 67.16 \\
    $L_2$-Global-MBP & 64.17 & 78.64 & 54.47 & 75.51 & 72.27 & 59.26 & 60.10 & 71.50 & 66.99 \\
    $L_2$-Gradient-MBP & 61.11 & 73.77 & 53.25 & 79.56 & 65.89 & 57.35 & 59.33 & 71.59 & 65.23 \\
    Lookahead & 60.84 & 79.18 & 54.44 & 71.05 & 68.76 & 55.94 & 53.41 & 71.26 & 64.36 \\
    LAMP & 58.04 & 63.64 & 51.92 & 66.05 & 67.43 & 55.36 & 52.42 & 71.09 & 60.74 \\
    \midrule
     \midrule
    Cosine-MBP & 66.20 & 79.15 & 55.62 & 78.45 & 71.62 & 57.56 & 61.37 & 72.51 & \underline{\textbf{67.81}} \\
    Frobenius-MBP & 65.71 & 79.84 & 55.61 & 78.78 & 71.62 & 61.62 & 61.37 &  71.48 & \underline{\textbf{68.25}}$^{\dagger}$ \\
    \bottomrule[1.05pt]
    \end{tabular}
    }
    \caption{\textbf{Overall XGLUE Score for Iterative Pruning of XLM-R$_{\mathrm{Base}}$ @ 31\% Remaining Weights.}}
    \label{tab:ft_all}}
\end{center}
\vspace{-4mm}
\end{table*}

\paragraph{Discussion}
\begin{figure}[ht]
\vspace{-3mm}
  \centering
  \includegraphics[width=1.\linewidth]{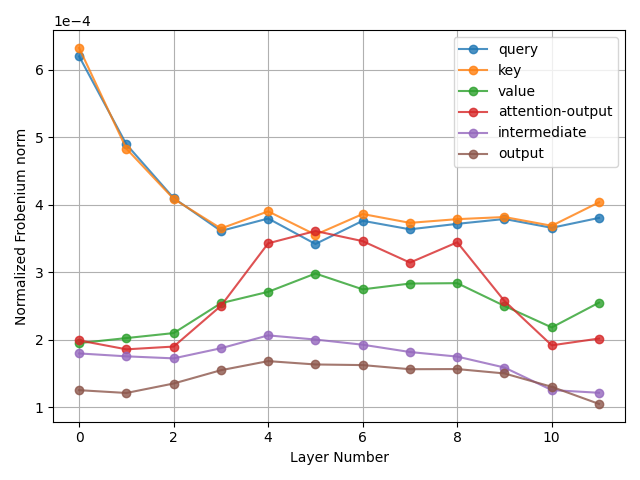}
   \vspace{-4mm}
\caption{\textbf{Pruned Model Weight Norms Per Layer}}\label{fig:prune_norm}
\end{figure}

From our experiments, we found that layer-wise pruning tends to outperform global pruning. This can be explained by the clear discrepancy between weight norms of different layer types within each self-attention block. Global pruning chooses the majority of weights to prune from the layer type that has the smallest norm, leading to an information bottleneck, or layer collapse~\cite{lee2018snip} for very high compression rates. This effect is due to layer normalization being applied after query, key and value (QKV) parameters, rescaling features such that weight magnitudes remain low. Hence, this motivates why we have focused on the application of {\em AlignReg} to layer-wise MBP. 
This is reflected in Figure~\ref{fig:prune_norm} which shows the weight norm by layer type for each layer for MBP. We see that QKV weight values are distinctly higher than the remaining fully-connected layers (attention output layer, intermediate position-wise feedforward layer and the blocks output layer), with the exception that the output attention layer norm becomes higher between layer 3-8.  

%
\begin{figure}[ht]
    \centering
    \includegraphics[scale=0.38]{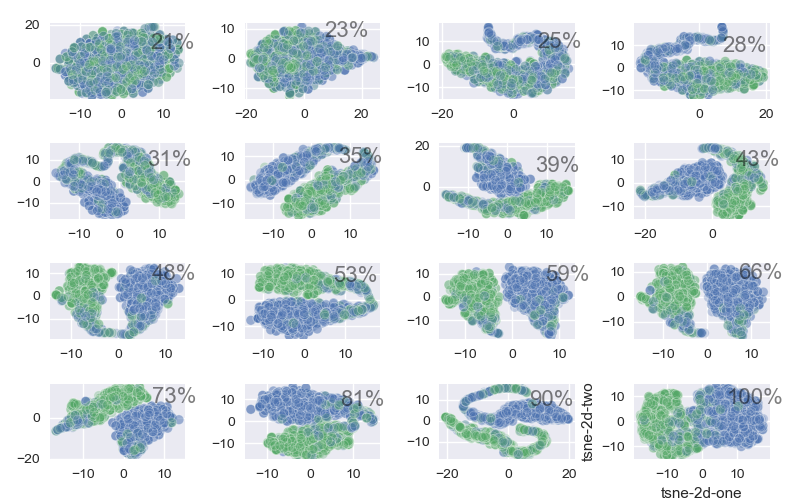}
    \vspace{-4mm}
    \caption{\textbf{Class Separability Between Class Representations At Each Iterative Pruning Step on PAWSX.}}
    \label{fig:tsne_viz}
\end{figure}

For the majority of tasks, the rate of performance drop for zero-shot test performance occurs close to 30\% of remaining weights. This is consistent for all pruning methods and therefore the focus of our analysis has been around this operating region. 

We also note that the effect of MBP (including our AlignReg regularization-based MBP) on zero-shot performance for different languages heavily depends on the semantic distance of evaluated language to the single language used for training. For example, in~\autoref{fig:xnli_perf} Arabic, Bulgarian, Swahili and Hindi have the largest drops in test accuracy around 20-60\% remaining weights. Similarly Arabic, Thai and Hindi suffer most around 20\% - 60\% for PoS tagging in~\autoref{fig:en_vs_avg_pos}. However, we also acknowledge this is partly reliant on the proportion of training data per language used during pretraining the underlying language model, in our case XLM-R$_{\mathrm{Base}}$.

Lastly, to show the representational degradation of pruned networks, in~\autoref{fig:tsne_viz} we visualize the class separability via a t-SNE plot of two principal components of the last hidden representation corresponding to the [CLS] token of an iteratively pruned XLM-R$_{\mathrm{Base}}$ for PAWSX. Even from only two principal components of a single token input, we clearly see a change in class separability from 31\% to 28\% remaining weights, reflecting the lack of linear separation.

\vspace{-2mm}
\section{Conclusion}
In this paper, we analysed iterative pruning in the zero-shot setting where a pretrained masked language model uses self-supervised learning on text from various languages but can only use a single language for downstream task fine-tuning. We find that some languages degrade in iterative pruning performance faster than others for some tasks (NER and XNLI) and propose a weight regularizer that biases the iteratively pruned model towards learning weight distributions close to the cross-lingually aligned pretrained state. This improves over well-established weight regularization methods for magnitude-based pruning in both the standard supervised learning setting and the zero-shot setting. 


\bibliography{anthology,custom}
\bibliographystyle{acl_natbib}
\end{document}